\documentclass{article} 
\usepackage{geolatent,times}

\usepackage{amsmath,amsfonts,bm}

\def\eqref#1{equation~\ref{#1}}

\def\1{\bm{1}}

\DeclareMathAlphabet{\mathsfit}{\encodingdefault}{\sfdefault}{m}{sl}
\SetMathAlphabet{\mathsfit}{bold}{\encodingdefault}{\sfdefault}{bx}{n}

\usepackage[hidelinks]{hyperref}
\usepackage{url}
\usepackage{amsmath,amssymb,amsthm}
\usepackage{mathtools}
\usepackage{booktabs}
\usepackage{multirow}
\usepackage{graphicx}
\usepackage{subcaption}
\usepackage{xcolor}
\usepackage{placeins}
\newtheorem{proposition}{Proposition}

\hypersetup{
  pdftitle={GeoLatent: Geometry-Guided Latent Structuring with Routed Optimization for 3D Reasoning},
  pdfauthor={Yakun Zhu, Yi Bin, Yujuan Ding, Zheng Wang, Pengpeng Zeng, Duo Peng, Jingkuan Song, Heng Tao Shen},
  pdfsubject={Preprint},
  pdfkeywords={3D reasoning, latent reasoning, vision-language models}
}

\title{GeoLatent: Geometry-Guided Latent\\
Structuring with Routed Optimization\\
for 3D Reasoning}

\author{%
  \textbf{Yakun Zhu\textsuperscript{1},\quad Yi Bin\textsuperscript{1,*},\quad Yujuan Ding\textsuperscript{2},\quad Zheng Wang\textsuperscript{1},\quad Pengpeng Zeng\textsuperscript{1}}\\
  \textbf{Duo Peng\textsuperscript{1},\quad Jingkuan Song\textsuperscript{1},\quad Heng Tao Shen\textsuperscript{1}}\\[2pt]
  \textsuperscript{1}Tongji University\quad
  \textsuperscript{2}The Hong Kong Polytechnic University\\
  \textsuperscript{*}Corresponding author: Yi Bin (\texttt{yi.bin@hotmail.com})
}

\begin{document}

\maketitle

\begin{abstract}
Despite progress in vision-language models, 3D spatial reasoning from 2D images remains challenging. Text-based methods describe intermediate geometry with discrete tokens, limiting fidelity for continuous spatial relations. Continuous latents offer richer representations, but a single latent type does not explicitly separate the cues needed across spatial tasks. Decomposed spatial latents address this by representing position, direction, and global geometry separately under geometric supervision. Yet the geometry representation can still collapse toward one dominant direction, and unrestricted attention can leave the latents underused during answer learning. We introduce \textbf{GeoLatent}, combining Common--Residual Geometry Alignment (CR-GEO) with routed optimization to structure the geometry states while promoting latent-mediated answer learning. CR-GEO separates shared from residual teacher geometry; routed optimization jointly trains geometry and language, temporarily directs visual answer learning through the latents, and restores full attention with geometry supervision. In controlled comparisons, CR-GEO raises geometry effective rank from $1.00$ to $3.87$, while blocking latent readout at the bottleneck lowers direction accuracy from $89.1\%$ to $25.8\%$ on 128 fixed questions. After recovery, the differentiated geometry representation and latent-mediated visual route remain available alongside direct image access. GeoLatent achieves $73.0\%$ on SPAR-Bench and $72.1\%$ on SPBench, outperforming previously reported methods on both.
\end{abstract}

\section{Introduction}
\label{sec:intro}

Vision-language models (VLMs) have achieved remarkable progress in visual understanding and reasoning, yet 3D spatial reasoning remains challenging \citep{zhang2025spar,li2026spatialladder}. Spatial reasoning tasks require models to infer geometric properties such as depth, distance, relative position, orientation, and scene layout from 2D observations. To improve these capabilities, existing studies have explored spatial instruction tuning, progressive training, and explicit 3D or 2.5D representations \citep{chen2024spatialvlm,li2026spatialladder,wu2025spatialmllm}. However, when intermediate reasoning is explicitly modeled, it is still commonly expressed as text-based chain-of-thought. Continuous geometric information must then be represented through discrete language or numerical tokens, which may lose fine-grained spatial detail and make the reasoning more susceptible to language priors \citep{geanchor2026}.

Continuous latent reasoning provides an alternative by representing intermediate reasoning in continuous hidden states rather than expressing every step as natural-language tokens \citep{shen2025codi,li2026livr}. For spatial reasoning, however, these latent representations need to capture task-relevant spatial cues at different levels of granularity: some tasks require precise object positions or relative directions, whereas others depend on more global scene geometry. With a single latent type, spatial cues with different functions and levels of granularity have to share the same form of latent representation, making it difficult to accommodate the diverse demands of spatial reasoning tasks. GeoAnchor addresses this limitation by decomposing intermediate 3D spatial information into three complementary latent components: position, direction, and geometry latents, which represent object positions, relative directions, and global scene structure, respectively \citep{geanchor2026}. These latent components are trained with corresponding geometric supervision. However, it remains unclear whether geometric supervision alone is sufficient to yield effective latent representations and whether these latent intermediates consistently contribute to subsequent reasoning and answer prediction.

Our analysis reveals two limitations that geometric supervision alone does not resolve. First, when the teacher geometry contains a dominant shared component, GeoAnchor's geometry alignment objective can lead multiple geometry states to learn highly similar representations rather than capture complementary scene structure. We observe that the resulting geometry representation has an effective rank close to one, indicating substantial redundancy across its states. Second, inserting decomposed spatial latents into the reasoning trajectory does not ensure that subsequent reasoning and answer prediction actually use them. Under standard multimodal attention, downstream text and answer tokens can still directly access the original image tokens, allowing the model to answer with limited reliance on the latent intermediates. Thus, geometric supervision alone does not ensure either differentiated geometry representations or effective use of the latent intermediates during answer learning.

\begin{figure}[t]
\centering
\includegraphics[width=\linewidth]{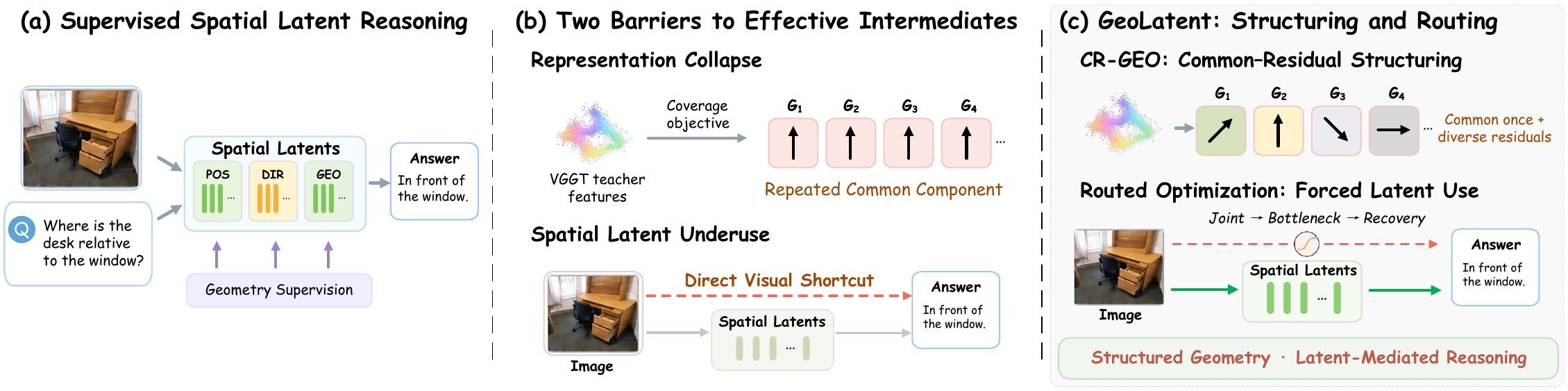}
\caption{GeoLatent addresses two limitations of decomposed latent reasoning: CR-GEO reduces redundancy among geometry states, while routed optimization encourages answer learning to use the decomposed spatial latents.}
\label{fig:motivation}
\end{figure}

As summarized in Figure~\ref{fig:motivation}, we introduce \textbf{GeoLatent}, which improves both the geometric representations learned by decomposed spatial latents and their role in subsequent spatial reasoning. Within GeoLatent, we first introduce \textbf{Common--Residual Geometry Alignment (CR-GEO)}, which models the geometry shared across teacher features once at the set level and encourages different geometry states to capture complementary residual geometry. This reduces redundant encoding and yields a more differentiated global geometry representation. We further develop \textbf{routed optimization}: training first jointly optimizes answer generation and geometric supervision, then uses a temporary visual bottleneck to direct answer-relevant visual information through the latent intermediates, and finally restores full attention while retaining geometric supervision. This allows the model to recover flexible multimodal reasoning while preserving the spatial representations learned during routed training.

Experimental results validate GeoLatent from multiple perspectives. Under identical initialization and batch order, replacing only GeoAnchor's geometry alignment objective with CR-GEO increases the effective rank of the geometry representation from $1.00$ to $3.87$. At the bottleneck stage, blocking downstream latent readout reduces accuracy on a fixed set of 128 direction questions from $89.1\%$ to $25.8\%$, showing that routed optimization makes answer prediction strongly dependent on the latent intermediates. After full visual access is restored, the differentiated geometry representation and the latent-mediated visual route remain available alongside direct image access; removing image content still causes substantial performance degradation, indicating that the final predictions remain strongly dependent on visual evidence. Overall, GeoLatent achieves $73.0\%$ on SPAR-Bench and $72.1\%$ on SPBench, improving over the reported GeoAnchor results by $4.6$ and $2.4$ points, respectively.

Our main contributions are summarized as follows:
\begin{itemize}
    \item We analyze decomposed spatial latent reasoning and identify two remaining limitations: GeoAnchor's geometry alignment can produce highly redundant global geometry representations, while latents inserted into the reasoning trajectory can remain underused during answer learning.
    
    \item We introduce \textbf{GeoLatent}, in which \textbf{CR-GEO} separates shared geometry from residual variation to learn more differentiated geometry representations, while \textbf{routed optimization} encourages answer-relevant visual information to pass through the latent intermediates during training.
    
    \item Representation analyses, intervention experiments, and benchmark evaluations show that GeoLatent learns more differentiated geometry representations, establishes strong dependence on latent intermediates during routed training, and further improves spatial reasoning performance.
\end{itemize}

\section{Related work}
\label{sec:related}

\subsection{Spatial reasoning in vision-language models}

Recent work improves spatial reasoning in VLMs mainly through spatial instruction tuning and progressive training \citep{chen2024spatialvlm,cheng2024spatialrgpt,li2026spatialladder}, or by incorporating explicit 2.5D/3D information such as depth, point clouds, and structured spatial representations \citep{wu2025spatialmllm,zhang2025spatialunderstanding}. These approaches provide stronger spatial supervision or geometric cues, but intermediate reasoning is still often expressed in discrete language, which can be insufficient for representing continuous spatial relations. More recently, GeoAnchor \citep{geanchor2026} moves beyond a uniform latent representation by decomposing 3D spatial information into complementary position, direction, and geometry latents with geometric supervision.

\subsection{Latent reasoning and visual intermediates}

Latent reasoning replaces verbalized intermediate steps with hidden representations. CODI \citep{shen2025codi} compresses explicit chain-of-thought reasoning into a continuous latent space through self-distillation, while pause-token methods introduce additional internal computation without explicit intermediate symbols \citep{goyal2024pause}. In multimodal reasoning, methods such as Mirage and LVR use latent visual intermediates tied to task-specific visual targets, while Aurora introduces predefined perception tokens for visual quantities such as depth and bounding boxes \citep{yang2026mirage,li2026lvr,bigverdi2025aurora}. LIVR \citep{li2026livr} instead learns latent visual tokens without explicit
intermediate supervision and uses visual bottlenecking to force task-relevant
visual information through these tokens before answer prediction. RIS \citep{cui2026ris} identifies latent-trajectory collapse and answer bypass
and addresses them with spatial-semantic grounding and progressive bottlenecking. Recent
mechanistic evidence shows that latent-reasoning gains can reflect boundary
markers, formatting, or attention patterns rather than slot content alone
\citep{guo2026beyond}. Accordingly, GeoLatent evaluates both the geometric
organization of decomposed spatial latents and their use during answer learning.

\subsection{Representation collapse}

Representation collapse has been widely studied in self-supervised representation learning \citep{caron2021dino,zbontar2021barlow,bardes2021vicreg}. Unlike the instance-agnostic latent-trajectory collapse studied in RIS \citep{cui2026ris}, our redundancy arises within an explicitly supervised set of geometry states under coverage-based alignment: balanced utilization does not ensure that different states encode complementary information. GeoLatent addresses this setting with CR-GEO, representing shared geometry at the set level while encouraging individual geometry states to capture residual variation. We use effective rank to quantify the resulting representation diversity \citep{roy2007effective}.

\section{Method}
\label{sec:method}

\begin{figure}[t]
\centering
\includegraphics[width=0.99\linewidth]{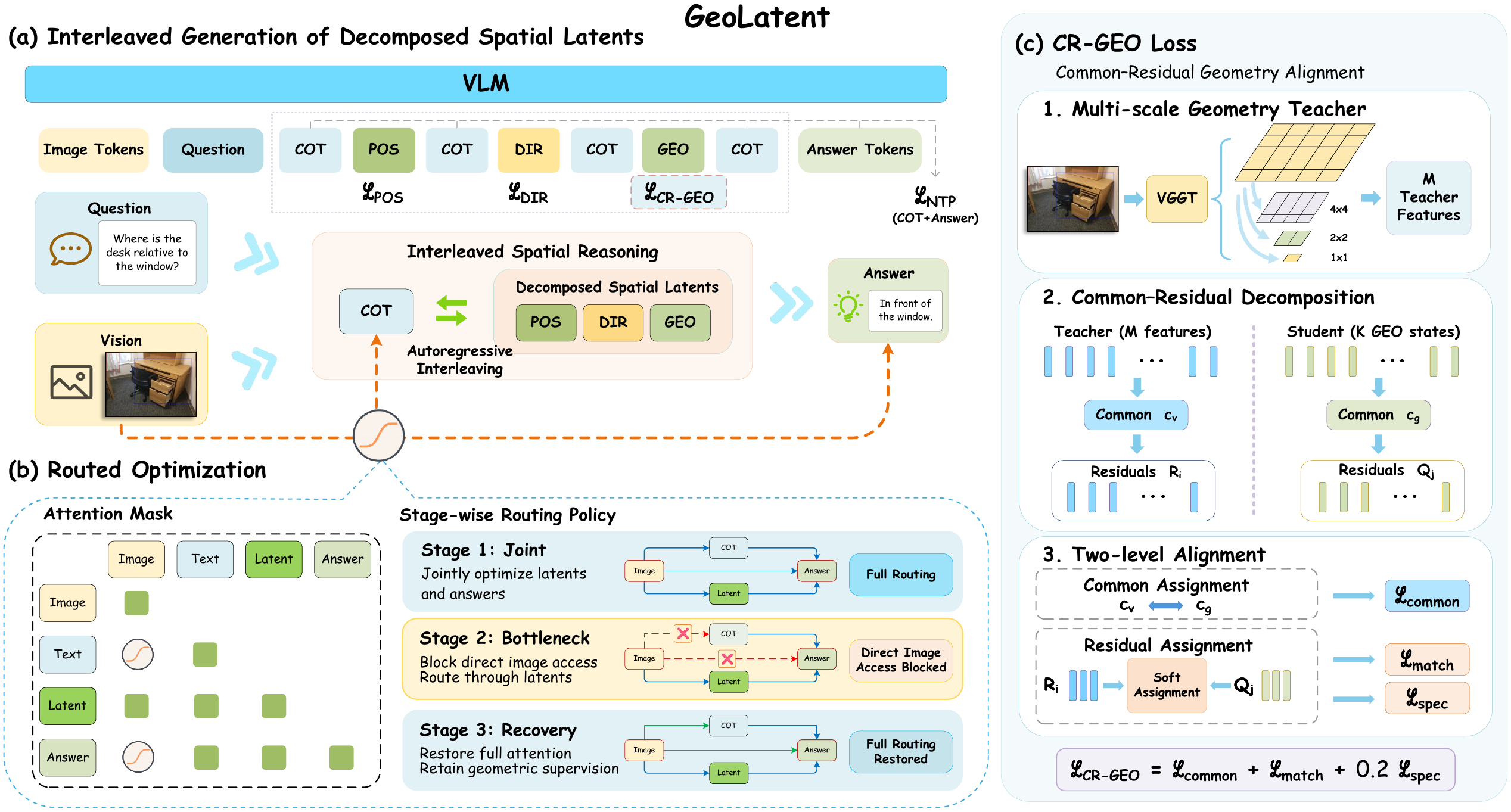}
\caption{GeoLatent overview. (a) A VLM interleaves language with decomposed POS, DIR, and GEO latents. (b) Routed optimization proceeds through joint, bottleneck, and recovery stages, temporarily restricting direct visual access during answer learning. (c) CR-GEO separates shared from residual teacher geometry for GEO supervision.}
\label{fig:architecture}
\end{figure}

GeoLatent builds on GeoAnchor's text--latent interleaved framework with decomposed position (POS), direction (DIR), and geometry (GEO) latents \citep{geanchor2026}. On top of this framework, we introduce two key designs: \textbf{Common--Residual Geometry Alignment (CR-GEO)} to improve the global geometry represented by the GEO latents, and \textbf{routed optimization} to promote the use of these latent intermediates during answer learning. Figure~\ref{fig:architecture} illustrates the overall framework.

\subsection{Preliminaries: Decomposed Spatial Latents}
\label{sec:prelim}

GeoLatent follows GeoAnchor's text--latent interleaved framework \citep{geanchor2026}, where language tokens and spatial latents are generated within the same autoregressive reasoning trajectory. Each latent consists of continuous hidden states that are mapped back to the language-model input space by a latent projector, allowing them to condition subsequent generation. Separate geometry heads map the latent hidden states to their corresponding geometric target spaces.

The decomposed latents encode complementary 3D information. POS and DIR represent local cues for object position and relative direction. Their hidden states are mean-pooled and mapped by separate linear heads to a camera-frame 3D position or unit direction vector, supervised with Smooth-$\ell_1$ \citep{girshick2015fastrcnn} and cosine losses, respectively. The corresponding targets are precomputed from Depth Anything v3 camera-frame back-projection \citep{lin2026depthanything3}. GEO instead represents global scene geometry through multiple latent states projected into the VGGT alignment space and supervised against multi-scale pooled VGGT features \citep{wang2025vggt}. Unlike POS and DIR, an individual GEO state is not assigned a predefined physical quantity; supervision is defined over the GEO set and the global teacher geometry.

For training stage $s$, the overall objective is
\begin{equation}
\mathcal{L}^{(s)}
=
\mathcal{L}_{\mathrm{NTP}}
+
\lambda_{\mathrm{local}}^{(s)}
\left(
\mathcal{L}_{\mathrm{pos}}
+
\mathcal{L}_{\mathrm{dir}}
\right)
+
\lambda_{\mathrm{GEO}}^{(s)}
\mathcal{L}_{\mathrm{GEO}},
\label{eq:overall}
\end{equation}
where $\mathcal{L}_{\mathrm{NTP}}$ supervises the reasoning text and final answer, $\mathcal{L}_{\mathrm{pos}}$ and $\mathcal{L}_{\mathrm{dir}}$ supervise the local spatial latents, and $\mathcal{L}_{\mathrm{GEO}}$ denotes the supervision objective for the global GEO latents. In this paper, we denote GeoAnchor's original coverage-based GEO objective as $\mathcal{L}_{\mathrm{cov}}$ and our proposed CR-GEO objective as $\mathcal{L}_{\mathrm{CR\text{-}GEO}}$. Stage-specific weights and implementation details are given in Appendix~\ref{app:training}. We next examine why the original GEO alignment can still yield highly redundant representations.

\subsection{Redundancy under Coverage-Based GEO Alignment}
\label{sec:collapse}

GeoAnchor aligns multiple GEO states with pooled VGGT teacher features using soft coverage and balanced utilization \citep{geanchor2026}. Let
$\hat{\mathbf v}_i$ denote an $\ell_2$-normalized teacher feature and
$\hat{\mathbf g}_j$ a projected, normalized GEO alignment feature. Define
\begin{equation}
A_{ij}
=
\frac{\hat{\mathbf v}_i^\top\hat{\mathbf g}_j}{\tau},
\qquad
p_{ij}
=
\operatorname{softmax}_{j}(A_{i\cdot}),
\qquad
u_j
=
\frac{1}{M}\sum_i p_{ij}.
\label{eq:geo_assignment}
\end{equation}
Ignoring numerical stabilizers, the coverage-based objective is
\begin{equation}
\mathcal{L}_{\mathrm{cov}}
=
-\frac{1}{M}\sum_i
\log\sum_j\exp(A_{ij})
+
\lambda_{\mathrm{bal}}
\sum_j u_j\log(Ku_j),
\label{eq:fuzzy}
\end{equation}
with $\tau=0.07$ and $\lambda_{\mathrm{bal}}=0.05$. The first term encourages coverage of the teacher features, while the second encourages balanced marginal use of the $K$ GEO states.

Balanced utilization, however, does not imply differentiated representations. If all GEO alignment features are identical, then $p_{ij}=1/K$ for all $i,j$, and consequently $u_j=1/K$, so the balance term vanishes up to numerical stabilization. Thus, all GEO states may repeatedly encode the same shared geometric component while fully satisfying the utilization constraint.

\begin{proposition}[Best collapsed configuration]
\label{prop:collapse}
Assume
$\left\|\frac{1}{M}\sum_i\hat{\mathbf v}_i\right\|_2>0$.
Among collapsed configurations
$\hat{\mathbf g}_1=\cdots=\hat{\mathbf g}_K=\mathbf g$
with $\|\mathbf g\|_2=1$, the optimum of
Equation~\ref{eq:fuzzy} is
\begin{equation}
\mathbf g^\star
=
\operatorname{normalize}
\left(
\frac{1}{M}\sum_i\hat{\mathbf v}_i
\right),
\qquad
\mathcal{L}_{\mathrm{cov}}(\mathbf g^\star)
=
-\log K-\frac{R_V}{\tau},
\end{equation}
where
$
R_V=
\left\|
\frac{1}{M}\sum_i\hat{\mathbf v}_i
\right\|_2
$.
\end{proposition}

Here, $\hat{\mathbf g}_j$ denotes the post-head, $\ell_2$-normalized GEO alignment feature. Appendix~\ref{app:proof} provides the proof. The proposition shows that when the teacher geometry contains a strong shared component, repeatedly aligning all GEO states with that component is explicitly favored within the collapsed family. The objective therefore does not require different GEO states to capture complementary residual geometry, motivating the common--residual formulation below.

\begin{figure}[t]
\centering
\includegraphics[width=0.78\linewidth]{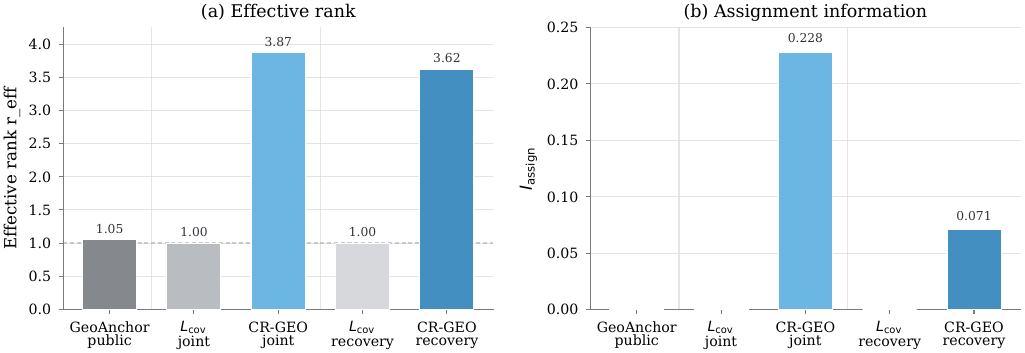}
\caption{GEO-token organization. The public checkpoint and models trained with the original GEO alignment loss approach rank one with near-zero assignment information, whereas CR-GEO yields differentiated GEO vectors. The joint pair comes from the matched-initialization loss-only comparison; the recovered pair comes from separately trained recovered controls in Table~\ref{tab:geo-ablation}.}
\label{fig:collapse}
\end{figure}

\subsection{Common--Residual Geometry Alignment}
\label{sec:crgeo}

CR-GEO separates the shared component of teacher geometry from its residual variation. Using the normalized teacher and GEO alignment features above, we define their common directions as
\begin{equation}
\mathbf c_v
=
\operatorname{normalize}
\left(
\frac{1}{M}\sum_i\hat{\mathbf v}_i
\right),
\qquad
\mathbf c_g
=
\operatorname{normalize}
\left(
\frac{1}{K}\sum_j\hat{\mathbf g}_j
\right).
\end{equation}
We then remove the teacher common direction from both teacher and student features:
\begin{equation}
\mathbf R_i
=
\hat{\mathbf v}_i
-
(\hat{\mathbf v}_i^\top\mathbf c_v)\mathbf c_v,
\qquad
\mathbf Q_j
=
\hat{\mathbf g}_j
-
(\hat{\mathbf g}_j^\top\mathbf c_v)\mathbf c_v.
\end{equation}
Let
$w_i=\|\mathbf R_i\|_2$,
and let
$\hat{\mathbf R}_i$ and $\hat{\mathbf Q}_j$
denote the normalized residuals. Residual-to-GEO assignments are
\begin{equation}
a_{ij}
=
\operatorname{softmax}_{j}
\left(
\frac{\hat{\mathbf R}_i^\top\hat{\mathbf Q}_j}{\tau_r}
\right),
\qquad
\tau_r=0.1.
\label{eq:res_assign}
\end{equation}

For our microbatch size of one, we write the objective in the following
per-example form and omit numerical stabilizers. With
$Z=\sum_i w_i$ and $u_j=Z^{-1}\sum_i w_i a_{ij}$,
CR-GEO is defined as
\begin{equation}
\begin{alignedat}{2}
\mathcal{L}_{\mathrm{common}}
&=
1-\mathbf c_g^\top\mathbf c_v,
&\qquad
\mathcal{L}_{\mathrm{match}}
&=
1-\frac{1}{Z}
\sum_{i,j}
w_i\,\mathrm{sg}(a_{ij})
\hat{\mathbf R}_i^\top\hat{\mathbf Q}_j,
\\
\bar H_{\mathrm{cond}}
&=
-\frac{1}{Z\log K}
\sum_{i,j}
w_i a_{ij}\log a_{ij},
&\qquad
\bar B_{\mathrm{bal}}
&=
\frac{1}{\log K}
\sum_j u_j\log(Ku_j),
\\
\mathcal{L}_{\mathrm{spec}}
&=
\bar H_{\mathrm{cond}}
+
0.1\bar B_{\mathrm{bal}},
&\qquad
\mathcal{L}_{\mathrm{CR\text{-}GEO}}
&=
\mathcal{L}_{\mathrm{common}}
+
\mathcal{L}_{\mathrm{match}}
+
0.2\mathcal{L}_{\mathrm{spec}}.
\end{alignedat}
\label{eq:crgeo}
\end{equation}
$\mathcal{L}_{\mathrm{common}}$ aligns the shared geometry once at the set level,
while $\mathcal{L}_{\mathrm{match}}$ and $\mathcal{L}_{\mathrm{spec}}$ encourage
different GEO states to capture complementary residual geometry. The assignments
are stop-gradient only in $\mathcal{L}_{\mathrm{match}}$.

\subsection{Routed Optimization}
\label{sec:pipeline}

Geometric supervision alone does not ensure that the resulting spatial latents are used during answer learning. Under standard multimodal attention, downstream text and answer tokens can directly access image features and may therefore rely only weakly on the latent intermediates. Inspired by the visual bottlenecking strategy of LIVR \citep{li2026livr}, we introduce a three-stage routed optimization schedule for decomposed spatial latents.

\textbf{Joint.}
We first jointly optimize next-token prediction and geometric supervision under standard multimodal attention. This allows the decomposed spatial latents to develop together with reasoning and answer generation.

\textbf{Bottleneck.}
We then temporarily prevent question, intermediate-text, and answer queries from attending directly to image keys, while latent queries retain visual access. Visual information needed by subsequent reasoning and answer prediction must therefore pass through the generated spatial latents. Geometric supervision is retained during this stage to preserve their spatial representations.

\textbf{Recovery.}
Finally, we restore full visual attention while continuing geometric supervision. The model can again combine direct image access with the latent-mediated route learned during bottleneck training, while retaining the differentiated geometric representations. Detailed routing masks, stage lengths, and loss weights are provided in Appendix~\ref{app:training}.

\section{Experiments}
\label{sec:exp}

We evaluate whether the decomposed latents are geometrically structured, used during answer learning, and visually dependent. Objective comparisons and POS/DIR decoding test representation structure; route interventions directly test latent use beyond task accuracy \citep{guo2026beyond}; benchmark and image interventions evaluate task performance and visual dependence.

\subsection{Experimental setup}
\label{sec:setup}

\textbf{Implementation.}
GeoLatent uses Qwen3-VL-2B-Instruct \citep{bai2025qwen3vl} with two hidden states for each POS/DIR latent and eight for GEO. Joint, bottleneck, and recovery run for 3{,}311 / 1{,}000 / 3{,}311 steps with learning rates $(2,1,2)\times10^{-5}$ and global batch 32. The vision encoder is frozen; the merger, language model, and latent modules are trainable. Controlled runs share the checkpoint, data, optimizer, and phase lengths; the loss-only comparison additionally matches initialization and per-rank batch order.

\textbf{Training data.}
We use 105{,}928 single-image samples: 100k released SPAR records \citep{zhang2025spar} from ScanNet, ScanNet++, and Structured3D \citep{dai2017scannet,yeshwanth2023scannetpp,zheng2020structured3d}, plus 5{,}928 single-image SpatialLadder samples \citep{li2026spatialladder}. POS/DIR targets use Depth Anything v3 camera-frame back-projection; GEO supervision uses VGGT features pooled at $\{1,2,4\}$.

\textbf{Evaluation.}
We evaluate on 2{,}866 single-image SPAR-Bench examples \citep{zhang2025spar}, 1{,}009 SPBench examples \citep{li2026spatialladder}, and the 4{,}607-example single-image subset of ViewSpatial \citep{li2026viewspatial}, following the official prompts and scorers. An overlap audit finds no exact input-image or image/frame--question--answer duplicates under the available metadata; excluding 37 records from two training-overlapping ScanNet scenes changes the three-benchmark mean only from $62.76$ to $62.59$. Appendix~\ref{app:details} provides full data construction, training, evaluation, and overlap-audit details.

\subsection{Main benchmark results}
\label{sec:main}

\begin{table}[t]
\centering
\caption{Spatial-reasoning performance (\%). $\dagger$: results reported by \citet{geanchor2026}; SPAR numerical entries use threshold-averaged accuracy; others use accuracy. Mean averages the three benchmarks. Best and second-best entries are bold and underlined, respectively.}
\label{tab:main}
\scriptsize
\setlength{\tabcolsep}{3.2pt}
\resizebox{\textwidth}{!}{%
\begin{tabular}{llccccccccccc}
\toprule
\multirow{2}{*}{Model} & \multirow{2}{*}{Params.} &
\multicolumn{6}{c}{SPAR-Bench} & \multicolumn{3}{c}{SPBench} &
ViewSpatial & \multirow{2}{*}{Mean} \\
\cmidrule(lr){3-8}\cmidrule(lr){9-11}\cmidrule(lr){12-12}
 & & Avg & Dep & Dis & Prox & Rel & View & Avg & Rel & Abs & Avg & \\
\midrule
\multicolumn{13}{l}{\emph{Proprietary models reported by \citet{geanchor2026}}} \\
GPT-4o$^\dagger$ \citep{hurst2024gpt4o} & -- & 40.1 & 34.3 & 43.4 & 57.1 & 45.9 & 31.0 & 53.4 & 49.4 & 56.0 & 37.5 & 43.7 \\
Gemini-2.5-Flash$^\dagger$ \citep{comanici2025gemini} & -- & 48.7 & 37.4 & 45.2 & 80.1 & 63.2 & 41.0 & 51.5 & 44.7 & 56.0 & 44.0 & 48.1 \\
\midrule
\multicolumn{13}{l}{\emph{General VLMs reported by \citet{geanchor2026}}} \\
MiniCPM-V-4.5$^\dagger$ \citep{yu2026minicpm} & 8B & 37.7 & 32.4 & 32.6 & 62.1 & 50.6 & 29.8 & 40.9 & 47.4 & 36.7 & 39.0 & 39.2 \\
LLaVA-OneVision-1.5$^\dagger$ \citep{an2025llavaov} & 8B & 34.5 & 29.3 & 32.6 & 58.5 & 40.7 & 26.7 & 40.8 & 45.1 & 38.0 & 38.7 & 38.0 \\
Qwen3-VL$^\dagger$ \citep{bai2025qwen3vl} & 8B & 41.1 & 32.5 & 34.6 & 71.2 & 62.1 & 31.5 & 55.1 & 47.6 & 60.0 & 44.0 & 46.7 \\
Molmo2$^\dagger$ \citep{clark2026molmo2} & 8B & 24.6 & 16.7 & 3.2 & 62.4 & 48.4 & 25.1 & 29.7 & 47.6 & 18.1 & 45.5 & 33.3 \\
GLM-4.1V$^\dagger$ \citep{hong2025glm41v} & 9B & 48.0 & 42.4 & 53.2 & 65.6 & 57.7 & 34.3 & 49.1 & 43.6 & 52.6 & 40.4 & 45.8 \\
Qwen3.5$^\dagger$ \citep{qwenteam2026qwen35} & 9B & 48.6 & 39.0 & 53.0 & 73.8 & 63.5 & 33.4 & 48.8 & 48.6 & 48.8 & \underline{45.6} & 47.7 \\
Kimi-VL$^\dagger$ \citep{kimitm2025kimivl} & 16B-A3B & 33.9 & 28.6 & 26.7 & 60.6 & 44.5 & 28.3 & 42.1 & 40.8 & 43.0 & 38.1 & 38.0 \\
InternVL3.5$^\dagger$ \citep{wang2025internvl35} & 38B & 38.0 & 31.5 & 35.5 & 60.9 & 58.0 & 25.5 & 53.1 & 49.4 & 55.5 & 41.4 & 44.2 \\
\midrule
\multicolumn{13}{l}{\emph{Specialized spatial models}} \\
G2VLM$^\dagger$ \citep{hu2026g2vlm} & 2B & 41.7 & 41.8 & 66.3 & 37.7 & 26.4 & 24.2 & 14.3 & 25.2 & 7.2 & 18.3 & 24.8 \\
3DThinker$^\dagger$ \citep{chen2026think3d} & 3B & 23.8 & 15.8 & 18.6 & 49.7 & 24.2 & 25.1 & 13.8 & 32.2 & 1.7 & 29.2 & 22.3 \\
SpatialLadder \citep{li2026spatialladder} & 3B & 31.4 & 20.6 & 28.1 & 55.0 & 41.2 & 29.8 & 66.7 & 73.3 & \underline{62.5} & 44.8 & 47.7 \\
SpatialMLLM-v1.1$^\dagger$ \citep{wu2025spatialmllm} & 4B & 33.7 & 25.3 & 34.2 & 53.8 & 36.5 & 31.2 & 54.2 & 55.7 & 53.3 & 39.7 & 42.5 \\
ViLaSR$^\dagger$ \citep{wu2025vilasr} & 7B & 38.6 & 28.9 & 43.1 & 61.8 & 46.4 & 28.2 & 52.2 & 54.4 & 50.8 & 34.3 & 41.7 \\
Aurora$^\dagger$ \citep{bigverdi2025aurora} & 13B & 29.6 & 26.9 & 29.0 & 52.7 & 21.7 & 25.7 & 32.1 & 36.8 & 29.0 & 25.6 & 29.1 \\
\midrule
\multicolumn{13}{l}{\emph{Base model and GeoAnchor references}} \\
Qwen3-VL-2B (base)$^\dagger$ & 2B & 32.4 & 21.9 & 24.0 & 60.6 & 49.2 & 28.3 & 52.9 & 51.4 & 54.8 & 36.3 & 40.5 \\
GeoAnchor$^\dagger$ & 2B & \underline{68.4} & \underline{55.5} & \underline{67.4} & \underline{80.3} & \underline{84.6} & \underline{68.8} & \underline{69.7} & \textbf{86.7} & 58.7 & \textbf{47.0} & \underline{61.7} \\
\midrule
\multicolumn{13}{l}{\emph{GeoLatent}} \\
\textbf{GeoLatent} & 2B & \textbf{73.0} & \textbf{58.3} & \textbf{72.3} &
\textbf{84.4} & \textbf{87.1} & \textbf{76.3} & \textbf{72.1} & \underline{84.1} &
\textbf{64.3} & 43.2 & \textbf{62.8} \\
\bottomrule
\end{tabular}%
}
\end{table}

\textbf{Task performance and reference-frame transfer.}
Table~\ref{tab:main} shows that GeoLatent reaches $73.0\%$ on SPAR-Bench and $72.1\%$ on SPBench, improving over the reported GeoAnchor results by $4.6$ and $2.4$ points. All five SPAR categories improve, with the largest gain on View ($+7.5$ points), while SPBench absolute spatial reasoning improves by $5.6$ points. The matched no-GEO control reaches a comparable three-benchmark mean ($62.4\%$ versus $62.8\%$), showing that aggregate accuracy alone cannot determine whether the representation is differentiated or the latent intermediates are used. On ViewSpatial, GeoLatent scores $43.2\%$, improves camera-relative direction by $5.9$ points over the same-data no-GEO control, and achieves the highest Camera score among the matched recovered controls ($53.2\%$ versus $48.7$--$49.4\%$). Person-centered questions additionally require estimating the reference person's orientation, which is not explicitly represented by the current latent vocabulary and remains a natural extension. The near-tie in aggregate accuracy therefore motivates separate analyses of representation structure, latent use, and visual dependence.

\subsection{Structuring the learned spatial representation}
\label{sec:ablation2}

We assess global GEO differentiation using effective rank $r_{\mathrm{eff}}$ and assignment information $I_{\mathrm{assign}}$: rank near one and $I_{\mathrm{assign}}{\approx}0$ indicate highly redundant states with nearly identical teacher assignments.

\begin{table}[t]
\centering
\caption{Core objective and routed-optimization controls. Left: GEO objectives at joint and recovered endpoints; right: a $2\times2$ post-joint comparison of bottleneck training and geometric supervision during recovery. Recovered controls match data, phase lengths, routing, and stage-specific local/GEO supervision weights; the original-objective control therefore uses GeoLatent's stage-specific weights. Mean is the unweighted average of SPAR-Bench, SPBench, and ViewSpatial.}
\label{tab:geo-ablation}
\normalsize
\setlength{\tabcolsep}{4pt}
\begin{minipage}[t]{0.60\linewidth}
\centering
\textit{(a) GEO objective and endpoint}\\
\begin{tabular*}{\linewidth}{@{\extracolsep{\fill}}llrrr@{}}
\toprule
Stage & Objective & Mean & $I_{\mathrm{assign}}$ & $r_{\mathrm{eff}}$ \\
\midrule
Joint & original $\mathcal{L}_{\mathrm{cov}}$ & 60.0 & $\approx0$ & 1.00 \\
Joint & \textbf{CR-GEO} & \textbf{61.3} & \textbf{.228} & \textbf{3.87} \\
Recovered & original $\mathcal{L}_{\mathrm{cov}}$ & 62.3 & $\approx0$ & 1.00 \\
Recovered & \textbf{CR-GEO} & \textbf{62.8} & \textbf{.071} & \textbf{3.62} \\
\bottomrule
\end{tabular*}
\end{minipage}\hfill
\begin{minipage}[t]{0.36\linewidth}
\centering
\textit{(b) Post-joint route}\\
\begin{tabular*}{\linewidth}{@{\extracolsep{\fill}}lr@{}}
\toprule
Route & Mean \\
\midrule
Direct NTP & 60.0 \\
Direct recovery & 61.2 \\
Bneck$\rightarrow$NTP & 60.2 \\
Bneck$\rightarrow$recovery & \textbf{62.8} \\
\bottomrule
\end{tabular*}
\end{minipage}
\end{table}

\textbf{Global GEO organization.}
Figure~\ref{fig:collapse} and Table~\ref{tab:geo-ablation}(a) show the core CR-GEO comparison. The matched joint runs start from bit-identical initialization, see the same per-rank batch sequence, and differ only in the GEO objective. The original $\mathcal{L}_{\mathrm{cov}}$ yields $I_{\mathrm{assign}}{\approx}0$ and $r_{\mathrm{eff}}{=}1.00$, whereas CR-GEO reaches $0.228$ and $3.87$. This structural change is accompanied by a $1.29$-point three-benchmark gain, including a $1.85$-point ViewSpatial gain (paired McNemar $p{=}0.016$). Importantly, the matched objective replacement improves all three benchmarks, so the increased representation diversity is accompanied by task gains rather than occurring only as a diagnostic change. At the separately trained recovered endpoints, the original objective remains rank one, while GeoLatent retains a differentiated representation with $I_{\mathrm{assign}}{=}0.071$ and $r_{\mathrm{eff}}{=}3.62$, close to the teacher-residual effective rank of approximately $4.1$.

\textbf{Local POS/DIR geometry.}
The final POS/DIR latents remain quantitatively decodable: on 504 held-out records from 248 training-disjoint scenes, GeoLatent obtains $0.338$\,m POS error and $0.899$ independent-DIR cosine. Replacing only the RGB input with a different-scene image raises POS error from $0.389$ to $1.474$\,m, showing that the decoded local geometry depends on the corresponding image.

\textbf{Geometry supervision during routed optimization.}
Table~\ref{tab:geo-ablation}(b) forms a $2\times2$ comparison of bottleneck training and retained geometric supervision. Retaining geometric supervision improves the three-benchmark mean by $1.2$ points without bottleneck training and by $2.6$ points after bottleneck training, yielding a positive interaction of $1.4$ points. The decomposed latent system can therefore retain geometric structure while supporting answer-relevant computation.

\subsection{Operational use and visual dependence}
\label{sec:intervention}

\begin{figure}[t]
\centering
\includegraphics[width=\linewidth]{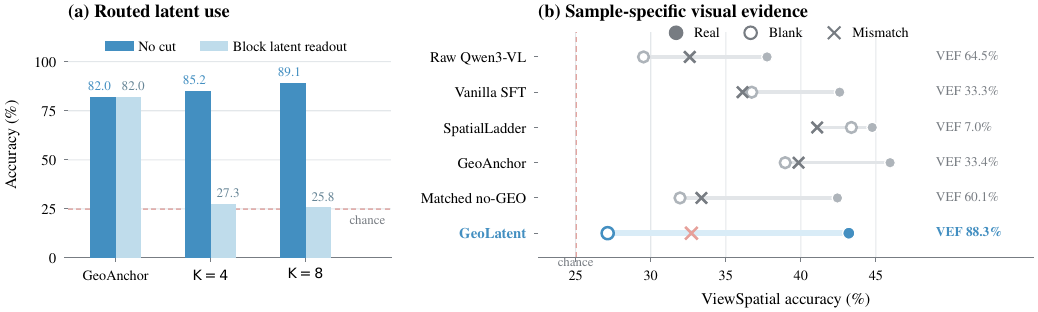}
\caption{Latent use and sample-specific visual dependence. (a) Blocking downstream readout from
all decomposed spatial latents leaves public GeoAnchor unchanged but reduces
the matched $K{=}4/8$ bottleneck models to near-chance accuracy.
(b) ViewSpatial under real (filled), blank (open), and mismatched images
 (cross), including the matched no-GEO control; VEF is the fraction of
 above-chance accuracy removed by blanking.}
\label{fig:kv}
\end{figure}

\textbf{Bottleneck-stage latent use.}
Table~\ref{tab:route-evidence}(a) shows that the bottleneck model retains
a $58.0\%$ three-benchmark mean when non-latent answer computation has no
direct image access, improving by $3.6$ points over the corresponding
original-$\mathcal{L}_{\mathrm{cov}}$ reference. On the fixed 128-example SPBench direction set in Figure~\ref{fig:kv}(a), blocking downstream readout from the decomposed latents reduces the matched $K{=}8$ model from $89.1\%$ to $25.8\%$, while public GeoAnchor remains at $82.0\%$. The same intervention changes accuracy by only $0.78$ points at the joint checkpoint, by $63.28$ points at bottleneck, and again by $0.78$ points after recovery. Strong dependence on decomposed latent readout is therefore localized to the bottleneck stage and does not imply the same dependence after recovery.

\begin{table}[t]
\centering
\caption{Operational evidence for routed optimization (\%).
Panel (a) compares the bottleneck endpoints of the main GeoLatent curriculum
and the original-$\mathcal{L}_{\mathrm{cov}}$ reference curriculum, where
non-latent queries have no direct image access.
In panel (b), \emph{Latent-only} removes direct image access for non-latent
queries, \emph{Neither} additionally removes latent readout, and direct recovery
skips bottleneck training. All recovery-route conditions use the
intervention-compatible attention path; its full-attention control therefore
differs slightly from the standard decoding path in Table~\ref{tab:main}.}
\label{tab:route-evidence}

\normalsize
\textit{(a) Bottleneck endpoint comparison}\\[2pt]
\begin{tabular}{lrrrr}
\toprule
Configuration & SPAR & SPBench & ViewSpatial & Mean \\
\midrule
original $\mathcal{L}_{\mathrm{cov}}$ reference
    & 66.4 & 64.2 & 32.8 & 54.5 \\
\textbf{GeoLatent (CR-GEO)}
    & \textbf{66.9} & \textbf{67.8} & \textbf{39.3} & \textbf{58.0} \\
\bottomrule
\end{tabular}

\medskip
\textit{(b) Recovered route retention}\\[2pt]
\begin{tabular}{llrrr}
\toprule
Model & Route condition & SPAR & SPBench & ViewSpatial \\
\midrule
GeoLatent & Full attention & 73.24 & 72.07 & 43.28 \\
GeoLatent & Latent-only
    & \textbf{52.20} & \textbf{51.27} & \textbf{33.49} \\
GeoLatent & Neither & 46.89 & 47.55 & 31.73 \\
\midrule
Direct recovery & Latent-only & 45.46 & 43.12 & 28.91 \\
Direct recovery & Neither & 45.52 & 43.47 & 28.93 \\
\bottomrule
\end{tabular}
\end{table}

\textbf{Recovery route retention.}
Table~\ref{tab:route-evidence}(b) isolates the image$\rightarrow$latent$\rightarrow$downstream route after full attention is restored. GeoLatent retains $52.2/51.3/33.5\%$ on SPAR-Bench/SPBench/ViewSpatial under the latent-only route; removing latent readout lowers performance by $5.3/3.7/1.8$ points, whereas the matched direct-recovery control shows approximately no corresponding gain. Thus, bottleneck training leaves a usable latent-mediated route after recovery, without implying that it remains the dominant path under full attention.

\textbf{Visual grounding.}
In Figure~\ref{fig:kv}(b), blanking or mismatching the image reduces GeoLatent's ViewSpatial accuracy from $43.2\%$ to $27.1\%$ or $32.7\%$; no-GEO reaches $42.4\%$ with the real image. Blanking removes $88.3\%$ of GeoLatent's above-chance margin, versus $60.1\%$ for no-GEO. On the Person subgroup, it raises no-GEO from $31.9\%$ to $44.7\%$ but lowers GeoLatent from $28.1\%$ to $24.8\%$, near chance. Thus, similar aggregate scores mask different grounding: GeoLatent's useful margin depends more on the corresponding image than on the Person-template prior. Substantial dependence on the corresponding image is also observed on
SPAR-Bench and SPBench.

\textbf{More comprehensive analyses.}
Due to space constraints, full ablations, cross-backbone and GEO-capacity studies, latent diagnostics, route-localization controls, visual-prior analyses, and statistical details are provided in Appendix~\ref{app:details}.

\section{Conclusion}
\label{sec:conclusion}

GeoLatent addresses redundant GEO representations and underuse of decomposed spatial latents. CR-GEO separates shared from residual teacher geometry, while routed optimization temporarily directs visual answer learning through the latents before restoring full attention. Experiments show differentiated GEO representations, decodable image-specific POS/DIR latents, strong bottleneck-stage latent use, and a usable latent-mediated route after recovery. Image interventions further show that the recovered predictions remain tied to corresponding visual evidence rather than the ViewSpatial Person-template prior. Together, these results improve spatial reasoning while separating representation structure, latent use, and visual dependence.

\bibliography{geolatent}
\bibliographystyle{geolatent}

\clearpage
\appendix

\section*{Technical Appendices and Supplementary Material}

\section{Proof of Proposition~\ref{prop:collapse}}
\label{app:proof}

For any collapsed normalized alignment configuration $\hat{\mathbf g}_1=\cdots=\hat{\mathbf g}_K=\mathbf g$, the balance term vanishes and the remaining objective is $-\log K-\mathbf{g}\cdot(\frac{1}{M}\sum_i\hat{\mathbf{v}}_i)/\tau$. Define
\[
\overline{\mathbf v}
=
\operatorname{normalize}
\left(
\frac{1}{M}\sum_i \hat{\mathbf v}_i
\right).
\]
By Cauchy--Schwarz, the objective is minimized by $\mathbf{g}=\overline{\mathbf{v}}$. With $\hat{\mathbf g}^\star_j=\overline{\mathbf{v}}$ for all $j$, all cosine similarities are equal: $A_{ij}=\hat{\mathbf{v}}_i\cdot\overline{\mathbf{v}}/\tau$ for every $j$. Hence $\sum_j \exp(A_{ij}) = K\exp(\hat{\mathbf{v}}_i\cdot\overline{\mathbf{v}}/\tau)$ and
\[
-\frac{1}{M}\sum_i \log\sum_j \exp(A_{ij})
= -\log K - \frac{1}{M\tau}\sum_i \hat{\mathbf{v}}_i\cdot\overline{\mathbf{v}}
= -\log K - \frac{1}{\tau}\Big\|\frac{1}{M}\sum_i\hat{\mathbf{v}}_i\Big\|,
\]
where the last step uses the definition of $\overline{\mathbf{v}}$. The assignment of every teacher feature is uniform over the $K$ latents, so $u_j=1/K$ and the balance term vanishes. \hfill$\square$

Numerically, on the frozen evaluation frame used for the representation diagnostics, the absolute difference between the analytical and computed collapsed-configuration losses is $1.4\times10^{-6}$ on average. The collapsed configuration attains a lower loss than the learned latents on 100\% of samples for every checkpoint tested (collapsed minus learned: public release $-0.82$, original-$\mathcal{L}_{\mathrm{cov}}$ joint with reference weights $-0.78$, and original-$\mathcal{L}_{\mathrm{cov}}$ recovery with reference weights $-0.05$). The teacher-feature spherical $k$-means baseline is worse on 100\% of samples (teacher-feature spherical $k$-means minus collapsed: $+0.198$).

\section{Experimental details and additional results}
\label{app:details}

\subsection{Data and benchmark construction}
\label{app:data}

All training and evaluation samples are single-image examples. Local POS and DIR targets use the camera coordinate frame: Depth Anything v3 (DA3) depth is back-projected with camera intrinsics and robust median aggregation inside each marked region, and DIR targets are normalized differences between the corresponding positions. The same construction is applied to every source.

\paragraph{Spatial reasoning training mixture.}
Table~\ref{tab:train-data} summarizes the 105{,}928-sample training mixture used by the main GeoLatent runs. It follows the source composition of the GeoAnchor spatial reasoning data: SPAR-derived ScanNet, ScanNet++, and Structured3D samples plus SpatialLadder single-image samples. In the released SPAR records, Structured3D spatial-imagination questions use sentence-format answers; we retain that released format for both corresponding buckets.

\begin{table}[ht]
\centering
\caption{Spatial reasoning training mixture. ``Select'' denotes multiple-choice answers and ``numeric'' denotes fill-in numerical answers. Structured3D spatial-imagination is sentence-format in the released SPAR records.}
\label{tab:train-data}
\scriptsize
\setlength{\tabcolsep}{4pt}
\begin{tabular}{lllc}
\toprule
Source & Task group & Answer format & Count \\
\midrule
\multirow{8}{*}{ScanNet / SPAR} & depth\_prediction\_oc & numeric & 4{,}000 \\
 & depth\_prediction\_oo & numeric & 4{,}000 \\
 & distance\_prediction\_oc & numeric & 4{,}000 \\
 & distance\_prediction\_oo & numeric & 4{,}000 \\
 & distance\_infer\_center\_oo & select & 8{,}000 \\
 & obj\_spatial\_relation\_oo & select & 8{,}000 \\
 & spatial\_imagination\_oc & select & 4{,}000 \\
 & spatial\_imagination\_oo & select & 4{,}000 \\
\midrule
\multirow{8}{*}{ScanNet++ / SPAR} & depth\_prediction\_oc & numeric & 4{,}000 \\
 & depth\_prediction\_oo & numeric & 4{,}000 \\
 & distance\_prediction\_oc & numeric & 4{,}000 \\
 & distance\_prediction\_oo & numeric & 4{,}000 \\
 & distance\_infer\_center\_oo & select & 4{,}000 \\
 & obj\_spatial\_relation\_oo & select & 4{,}000 \\
 & spatial\_imagination\_oc & select & 4{,}000 \\
 & spatial\_imagination\_oo & select & 4{,}000 \\
\midrule
\multirow{6}{*}{Structured3D / SPAR} & depth\_prediction\_oc & numeric & 4{,}000 \\
 & depth\_prediction\_oo & numeric & 4{,}000 \\
 & distance\_prediction\_oc & numeric & 4{,}000 \\
 & distance\_prediction\_oo & numeric & 8{,}000 \\
 & spatial\_imagination\_oc & sentence & 4{,}000 \\
 & spatial\_imagination\_oo & sentence & 4{,}000 \\
\midrule
\multirow{4}{*}{SpatialLadder-26K} & relative direction & select & 2{,}253 \\
 & absolute distance & numeric & 1{,}127 \\
 & object size & numeric & 1{,}514 \\
 & relative distance & select & 1{,}034 \\
\midrule
Total & -- & -- & 105{,}928 \\
\bottomrule
\end{tabular}
\end{table}

\paragraph{Geometry supervision.}
In geometry-enabled configurations, records may contain local POS/DIR spans and a global GEO span. POS and DIR receive local geometric supervision when the corresponding objects are present, while the GEO span is aligned to VGGT teacher features. Following GeoAnchor, we use the final VGGT feature map and apply average pooling at resolutions $\{1,2,4\}$, giving $1^2+2^2+4^2=21$ teacher vectors per image. GeoAnchor's GEO loss aligns the eight vectors within the GEO token to these teacher features with soft coverage and balance; GeoLatent replaces this alignment term with CR-GEO while leaving the teacher features and pooled resolution unchanged.

\subsection{Benchmark definitions}
\label{app:benchmarks}

SPAR-Bench and SPBench are in-domain with respect to the spatial reasoning sources, while ViewSpatial serves as an external cross-dataset transfer benchmark with distinct examples and partially overlapping spatial task definitions. Table~\ref{tab:bench-defs} lists the task groups and sample counts used by the evaluation harness.

\begin{table}[ht]
\centering
\caption{Benchmark task groups and sample counts. SPAR-Bench follows the single-image subset used in GeoAnchor. ViewSpatial additionally tests transfer from camera-frame supervision to camera- and person-centered reference frames.}
\label{tab:bench-defs}
\scriptsize
\setlength{\tabcolsep}{4pt}
\begin{tabular}{lllc}
\toprule
Benchmark & Reported group & Underlying task(s) & Count \\
\midrule
\multirow{5}{*}{SPAR-Bench} & Dep. & depth\_prediction\_oc, depth\_prediction\_oo & 732 \\
 & Dis. & distance\_prediction\_oc, distance\_prediction\_oo & 756 \\
 & Prox. & distance\_infer\_center\_oo & 340 \\
 & Rel. & obj\_spatial\_relation & 364 \\
 & View & spatial\_imagination\_oc, spatial\_imagination\_oo & 674 \\
\midrule
\multirow{2}{*}{SPBench} & Rel. & object\_rel\_direction, object\_rel\_distance & 397 \\
 & Abs. & object\_abs\_distance, object\_size\_estimation & 612 \\
\midrule
\multirow{4}{*}{ViewSpatial} & Camera--relative direction & camera-view relative direction & 1{,}773 \\
 & Camera--object view & camera-view object orientation & 996 \\
 & Person--object view & person-centered object/person orientation & 996 \\
 & Person--relative direction & person-centered relative direction & 842 \\
\midrule
Total & SPAR-Bench / SPBench / ViewSpatial & -- & 2{,}866 / 1{,}009 / 4{,}607 \\
\bottomrule
\end{tabular}
\end{table}

For SPAR-Bench, multiple-choice questions are scored by exact option correctness, while numerical questions use the benchmark's threshold-averaged accuracy over distance/depth tolerances. SPBench reports relative and absolute spatial categories. ViewSpatial reports camera-perspective and person-perspective subgroups; the main paper uses its overall average, while Appendix analyses use the four subgroups to characterize how the learned camera-frame geometry transfers across reference frames.

\paragraph{Train--benchmark overlap audit.}
We compare the final 105{,}928-record training mixture against every evaluated example using record IDs, normalized questions, exact input-image SHA-256, and, where available, ScanNet scene/frame identity. Table~\ref{tab:overlap-audit} separates shared domains and templates from duplicated examples. No benchmark record shares an exact input image or an image/frame--question--answer composite with training. SPAR-Bench does not expose scene/frame metadata, so no scene-disjoint claim is made for that benchmark. Its exact image and composite checks remain zero. SPBench and ViewSpatial contain 23 and 14 records from two ScanNet scenes also present in the SPAR-derived training source; removing those records changes SPBench from $72.11$ to $71.74\%$, ViewSpatial from $43.20$ to $43.04\%$, and the GeoLatent three-benchmark mean from $62.76$ to $62.59\%$. Repeated question strings in SPAR-Bench and SPBench arise on different images and reflect their templated, in-domain task definitions rather than duplicated QA examples.

\begin{table}[ht]
\centering
\caption{Overlap of each evaluation set with the 105{,}928-record training mixture. Entries count affected benchmark records. ``Exact example'' requires matching image or scene/frame, normalized question, and answer. A dash means the benchmark does not expose that metadata.}
\label{tab:overlap-audit}
\scriptsize
\setlength{\tabcolsep}{4pt}
\begin{tabular}{lrrrrrr}
\toprule
Benchmark & Record ID & Scene & Scene/frame & Image SHA & Question & Exact example \\
\midrule
SPAR-Bench & 0 & -- & -- & 0 & 63 & 0 \\
SPBench & 0 & 23 & 5 & 0 & 777 & 0 \\
ViewSpatial & 0 & 14 & 2 & 0 & 0 & 0 \\
\bottomrule
\end{tabular}
\end{table}

\subsection{Full blank- and mismatched-image interventions}
\label{app:blank-full}

For the blank condition, we replace every benchmark image with a same-size mid-gray canvas. For the mismatched condition, each image is replaced according to one fixed donor mapping with a different natural image from the same benchmark and task/answer-format group. Donors always have a different image hash; they additionally come from a different scene whenever scene metadata is available. Questions, choices, decoding configuration, and scorer are unchanged within each comparison. Trained models require a complete \texttt{<answer>} span, whereas the raw instruct model uses the official flexible parser because it was not trained with this tagged protocol. ``Vanilla SFT'' is a same-data, answer-only reference without decomposed spatial latents, adapted using LoRA \citep{hu2021lora} ($r{=}32$, $\alpha{=}64$, learning rate $2\times10^{-5}$, two epochs). All runs use the complete fixed denominator.

\begin{table}[ht]
\centering
\caption{Visual-evidence interventions. Panel (a) reports full-benchmark accuracy/score as real/blank/mismatched. Panel (b) decomposes ViewSpatial into Camera and Person aggregates. Direct recovery skips the bottleneck; SpatialLadder uses matched pixel bounds.}
\label{tab:blank-full}
\scriptsize
\setlength{\tabcolsep}{4pt}
\textit{(a) Full benchmark matrix (real / blank / mismatched)}\\[2pt]
\begin{tabular*}{\linewidth}{@{\extracolsep{\fill}}lccc@{}}
\toprule
Model & SPAR-Bench & SPBench & ViewSpatial \\
\midrule
Raw Qwen3-VL & $28.97/20.83/26.36$ & $52.06/23.48/34.72$ & $37.75/29.52/32.60$ \\
Vanilla SFT & $62.01/41.94/39.44$ & $62.38/41.85/41.86$ & $42.59/36.73/36.12$ \\
SpatialLadder & $31.44/30.80/29.10$ & $66.75/45.95/43.86$ & $44.76/43.37/41.09$ \\
GeoAnchor & $67.66/44.23/39.13$ & $69.22/43.52/41.62$ & $45.95/38.96/39.85$ \\
Matched no-GEO & $72.92/47.45/38.34$ & $71.78/41.57/42.53$ & $42.44/31.95/33.38$ \\
Original $L_{\rm cov}$ & $73.21/45.90/38.21$ & $70.43/41.62/43.71$ & $43.22/27.41/34.17$ \\
Direct recovery & $72.08/45.62/38.20$ & $71.15/41.17/43.12$ & $40.42/26.89/30.61$ \\
\textbf{GeoLatent} & $\mathbf{72.98/46.03/38.07}$ & $\mathbf{72.11/43.82/44.53}$ & $\mathbf{43.20/27.13/32.71}$ \\
\bottomrule
\end{tabular*}

\vspace{5pt}
\textit{(b) ViewSpatial reference-frame aggregates}\\[2pt]
\begin{tabular*}{\linewidth}{@{\extracolsep{\fill}}lrrrrrr@{}}
\toprule
& \multicolumn{3}{c}{Camera} & \multicolumn{3}{c}{Person} \\
\cmidrule(lr){2-4}\cmidrule(lr){5-7}
Model & Real & Blank & Mismatch & Real & Blank & Mismatch \\
\midrule
Raw Qwen3-VL & 38.82 & 34.34 & 34.20 & 36.13 & 22.25 & 30.20 \\
Vanilla SFT & 37.92 & 30.01 & 27.16 & 49.62 & 46.84 & 49.62 \\
SpatialLadder & 38.35 & 31.53 & 29.97 & 54.41 & 61.21 & 57.83 \\
GeoAnchor & 45.14 & 31.17 & 33.80 & 47.17 & 50.71 & 48.97 \\
Matched no-GEO & 49.44 & 23.47 & 31.78 & 31.88 & 44.72 & 35.80 \\
Original $L_{\rm cov}$ & 48.72 & 22.32 & 31.02 & 34.93 & 35.09 & 38.90 \\
Direct recovery & 49.19 & 29.40 & 31.64 & 27.20 & 23.12 & 29.05 \\
\textbf{GeoLatent} & \textbf{53.20} & \textbf{28.67} & 34.16 & 28.13 & \textbf{24.81} & 30.52 \\
\bottomrule
\end{tabular*}
\end{table}

Visual dependence is not confined to ViewSpatial. Matched no-GEO loses $25.5/30.2$ points under blanking and $34.6/29.3$ under mismatched images on SPAR-Bench/SPBench. The control using GeoAnchor's GEO loss and the direct-recovery control fall from $62.28/61.22\%$ real-image three-benchmark means to $38.31/37.89\%$ when blanked and $38.69/37.31\%$ when mismatched. GeoLatent loses $27.0/28.3$ points under blanking on SPAR-Bench/SPBench and $34.9/27.6/10.5$ under mismatched images on SPAR-Bench/SPBench/ViewSpatial; for each benchmark, no sampled sign flip is as extreme as observed (100{,}000 draws; plus-one $p{=}1/100001$). Thus the closely scoring recovered controls remain strongly dependent on visual input; GeoLatent combines this evidence dependence with a differentiated GEO representation and the highest ViewSpatial Camera score among the evaluated models and controls.

SpatialLadder exhibits a different benchmark-dependent pattern: real/blank/mismatched changes only from $31.44/30.80/29.10\%$ on SPAR-Bench, but from $66.75/45.95/43.86\%$ on SPBench. On ViewSpatial it loses only $3.7$ points under mismatched images overall ($44.76\to41.09$), with the loss concentrated in Camera questions while Person accuracy remains above its real-image score.

The reference-frame decomposition localizes ViewSpatial's language prior. Vanilla SFT loses only $2.78$ points on Person questions, while SpatialLadder, GeoAnchor, and matched no-GEO improve after blanking; no-GEO separates sharply into Camera $49.44\to23.47\%$ and Person $31.88\to44.72\%$. Thus its strong in-domain accuracy is visually supported, but part of its ViewSpatial Person score remains available without the corresponding image. SpatialLadder's Camera questions show the opposite pattern: real $38.35$ versus $31.53$ blank and $29.97$ mismatched. In contrast, GeoLatent's blank Person score is $24.81\%$, effectively the four-way chance level. Its real-image strength is concentrated in Camera questions, where it reaches $53.20\%$---$8.1$ points above GeoAnchor---and falls by $24.53$ points when the image is removed. Thus GeoLatent's ViewSpatial predictions derive their useful margin from the visually grounded reference frame rather than the Person-template prior. The following audit connects this behavior to the benchmark's question and answer distributions.

\subsection{Template and answer priors in ViewSpatial}
\label{app:visual-priors}

We audit the 20{,}680 spatial QA records in SpatialLadder-26K against the 4{,}607 ViewSpatial questions used in our single-image evaluation. SpatialLadder contains 2{,}253 single-image camera-relative-direction questions of the form ``From the camera's perspective, is object A to object B's left, right, front, or back?'' ViewSpatial uses surface paraphrases of the same relation-and-reference-frame contract for its 1{,}773 Camera--Relative-Direction questions. SpatialLadder also contains 2{,}162 multi-view questions of the form ``standing by A and facing B, where is C?'', closely matching ViewSpatial's scene-simulation family. The latter family is outside our 4{,}607-question four-task subset, but demonstrates that the two resources share a common template vocabulary and reference-frame construction.

After lowercase and punctuation normalization, no complete question string is shared between the 20{,}680 SpatialLadder records and the 4{,}607 ViewSpatial questions. The exploitable signal is therefore a task-distribution prior rather than verbatim QA duplication. Table~\ref{tab:view-priors} separates answer-position balance from answer-semantics balance.

\begin{table}[ht]
\centering
\caption{ViewSpatial answer priors. ``Letter max'' is the frequency of the most common correct option position; ``semantic max'' is the most frequent direction irrespective of option position.}
\label{tab:view-priors}
\scriptsize
\setlength{\tabcolsep}{4pt}
\begin{tabular}{lrrrr}
\toprule
Task & $N$ & Letter max & Semantic max & Direction \\
\midrule
Camera--relative direction & 1{,}773 & 26.4 & 16.0 & right \\
Camera--object orientation & 996 & 26.0 & 26.8 & back \\
Person--object orientation & 996 & 26.1 & \textbf{57.2} & front \\
Person--relative direction & 842 & 34.8 & 27.4 & right \\
\bottomrule
\end{tabular}
\end{table}

The strongest prior occurs in Person--Object-Orientation: 570 of 996 gold answers are semantically ``front,'' although their A--D positions remain balanced. With blank images, SpatialLadder predicts ``front'' on 614 examples, ``front-left'' on 241, and ``front-right'' on 81, reaching $76.10\%$ on this subgroup and $74.20\%$ with mismatched images. This behavior is consistent with reliance on direction semantics and template-conditioned option structure rather than a fixed answer letter. Its overall accuracy is $43.37\%$ with blank images and $41.09\%$ with mismatched images, compared with $44.76\%$ on real images.

GeoLatent provides an informative contrast because its training mixture includes all 5{,}928 SpatialLadder single-image examples, including the 2{,}253 camera-relative questions. Nevertheless, blanking ViewSpatial images reduces GeoLatent from $43.20\%$ to $27.13\%$, close to the $25\%$ chance level. Exposure to the template family therefore does not by itself account for GeoLatent's normal-image performance: its recovered predictions derive most of their above-chance ViewSpatial accuracy from visual evidence.

\subsection{Training configuration}
\label{app:training}

Table~\ref{tab:train-config} summarizes the three training phases and their stage-specific objective weights.
\begin{table}[ht]
\centering
\caption{GeoLatent training phases. All phases use Qwen3-VL-2B-Instruct, frozen vision encoder, trainable merger/LLM/latent modules, global batch 32, bf16 computation, and fp32 optimizer states.}
\label{tab:train-config}
\small
\begin{tabular}{lcccccc}
\toprule
Phase & Steps & LR & Attention policy & $\lambda_{\mathrm{NTP}}$ & $\lambda_{\mathrm{local}}$ & $\lambda_{\mathrm{GEO}}$ \\
\midrule
Joint & 3{,}311 & $2\times10^{-5}$ & full & 1.0 & 1.0 & 0.2 \\
Bottleneck & 1{,}000 & $1\times10^{-5}$ & decomposed-latent bottleneck & 1.0 & 0.1 & 0.1 \\
Recovery & 3{,}311 & $2\times10^{-5}$ & full & 1.0 & 0.1 & 0.2 \\
\bottomrule
\end{tabular}
\end{table}

\paragraph{Reference configuration with original $\mathcal{L}_{\mathrm{cov}}$.}
The reference configuration shares the data, Qwen3-VL-2B-Instruct initialization, phase lengths, and learning rates, with local/GEO supervision weights $1.0/0.1$ in joint training, $0.1/0.01$ in bottleneck training, and $0.01/0.01$ in recovery. GeoLatent uses $1.0/0.2$, $0.1/0.1$, and $0.1/0.2$, respectively. These rows compare full training curricula; the matched-initialization experiment below isolates the GEO loss itself.

\paragraph{Matched-initialization objective test.}
For the loss-only joint test, we construct one shared seed-3407 initialization after all added latent modules are instantiated, then launch both arms from those bit-identical tensors. The per-rank batch sequences, data, optimizer, learning rate, step count, attention policy, and stage-specific local/GEO supervision weights are identical; the only changed field is $\mathcal{L}_{\mathrm{CR\text{-}GEO}}$ versus $\mathcal{L}_{\mathrm{cov}}$. The shared-init fingerprint and the first 64 batches on every rank match exactly. These are the loss-only joint rows of Table~\ref{tab:geo-ablation}. The recovered endpoints match data, phase lengths, routing schedule, and GeoLatent's stage-specific local/GEO supervision weights, but are trained separately under different GEO objectives; they are not continuations of the shared-initialization joint pair.

During bottleneck training, question, intermediate-text, and answer queries cannot directly attend to image keys, while the decomposed latent queries retain image access. Recovery restores full attention while continuing the corresponding local and GEO supervision. These controls isolate how routed answer learning and retained geometric supervision affect the learned latent representations.

\subsection{Full objective and curriculum ablation}
\label{app:full-ablation}

All rows in Table~\ref{tab:ablation-full} use the same 105{,}928 samples, Qwen3-VL-2B-Instruct base checkpoint, and initialization procedure unless otherwise noted.
\begin{table}[ht]
\centering
\caption{Full loss and curriculum matrix. The post-joint recovery variants use local/GEO supervision weights $0/0$ (NTP-only), $0.1/0$ (local-only), $0.01/0.05$, and $0.1/0.2$ (standard GeoLatent recovery). ``Reference weights'' denotes the stage-specific weights of the original-$\mathcal{L}_{\mathrm{cov}}$ reference configuration, while ``GeoLatent weights'' denotes GeoLatent's stage-specific local/GEO supervision weights.}
\label{tab:ablation-full}
\scriptsize
\resizebox{\linewidth}{!}{%
\begin{tabular}{llcccc}
\toprule
GEO objective & Training route & SPAR & SPBench & ViewSpatial & Mean \\
\midrule
\multicolumn{6}{l}{\emph{Matched-initialization loss-only joint test}} \\
original GEO $\mathcal{L}_{\mathrm{cov}}$ & joint only & 68.6 & 70.1 & 41.3 & 60.0 \\
\textbf{CR-GEO $\mathcal{L}_{\mathrm{CR\text{-}GEO}}$} & \textbf{joint only} & \textbf{69.6} & \textbf{71.2} & \textbf{43.1} & \textbf{61.3} \\
\midrule
\multicolumn{6}{l}{\emph{Additional independently trained joint checkpoints}} \\
no GEO span/objective & joint only & 68.4 & 69.4 & 43.0 & 60.3 \\
original GEO $\mathcal{L}_{\mathrm{cov}}$ (reference weights) & joint only & 68.3 & 69.9 & 40.5 & 59.6 \\
original GEO $\mathcal{L}_{\mathrm{cov}}$ (GeoLatent weights; independent init) & joint only & 67.9 & 63.6 & 36.9 & 56.1 \\
CR-GEO $\mathcal{L}_{\mathrm{CR\text{-}GEO}}$ (independent init) & joint only & 68.1 & 70.4 & 44.2 & 60.9 \\
\midrule
\multicolumn{6}{l}{\emph{Post-joint controls}} \\
no GEO span/objective & bottleneck $\rightarrow$ recovery & 72.9 & 71.8 & 42.4 & 62.4 \\
original GEO $\mathcal{L}_{\mathrm{cov}}$ (reference weights) & bottleneck $\rightarrow$ recovery & 73.3 & 71.3 & 42.2 & 62.2 \\
original GEO $\mathcal{L}_{\mathrm{cov}}$ (GeoLatent weights) & bottleneck $\rightarrow$ recovery & 73.2 & 70.4 & 43.2 & 62.3 \\
original GEO $\mathcal{L}_{\mathrm{cov}}$ & direct NTP continuation & 72.4 & 72.0 & 44.2 & 62.9 \\
CR-GEO & direct NTP continuation & 72.8 & 70.7 & 36.6 & 60.0 \\
CR-GEO & direct recovery (no bottleneck) & 72.1 & 71.2 & 40.4 & 61.2 \\
CR-GEO & bottleneck $\rightarrow$ NTP-only recovery & 71.6 & 70.8 & 38.2 & 60.2 \\
CR-GEO & bottleneck $\rightarrow$ local-only recovery & 72.4 & 71.5 & 41.8 & 61.9 \\
CR-GEO & bottleneck $\rightarrow$ recovery ($0.01/0.05$) & 73.3 & 71.7 & 40.3 & 61.8 \\
\textbf{CR-GEO} & \textbf{bottleneck $\rightarrow$ recovery} & 73.0 & \textbf{72.1} & 43.2 & \textbf{62.8} \\
\bottomrule
\end{tabular}%
}
\end{table}

Table~\ref{tab:route-evidence}(a) reports the bottleneck endpoints in the main text. GeoLatent preserves strong in-domain spatial reasoning while operating through the decomposed latents and improves the three-benchmark mean by approximately $3.6$ points over the corresponding model using GeoAnchor's GEO loss. This end-to-end stage comparison complements the matched-initialization joint test, which isolates the GEO loss itself.

Local-only recovery and recovery with local/GEO weights $0.01/0.05$ improve over bottleneck-to-NTP recovery by $1.5$--$1.7$ points, while the standard GeoLatent recovery weights $0.1/0.2$ yield the largest gain. The full-curriculum controls further show that recovery can restore task performance under the original objective, whereas in this comparison only the CR-GEO configuration builds and retains a differentiated GEO representation.

The matched-initialization joint rows isolate the GEO-loss effect. For these fixed checkpoints, CR-GEO improves SPAR-Bench, SPBench, and ViewSpatial by $0.9$, $1.1$, and $1.85$ points, respectively, for a $1.29$-point gain in their equal-weight mean; the paired ViewSpatial difference is $+1.85$ points (McNemar $p{=}0.016$, bootstrap 95\% CI $[+0.35,+3.34]$). The same starting tensors diverge to $(I_{\mathrm{assign}},r_{\mathrm{eff}})=(0.228,3.87)$ under CR-GEO and approximately $(0,1.00)$ under GeoAnchor's GEO loss. The separately trained full-curriculum endpoints show the same structural distinction: the original-objective control using GeoLatent weights remains rank one, while GeoLatent reaches $(0.071,3.62)$ after bottleneck and recovery.

\paragraph{GEO-span capacity.}
Table~\ref{tab:geo-count-full} varies only the number of vectors in the GEO token. The four runs share the same backbone initialization, data and batch order, random seed, objective weights, routing schedule, and optimization hyperparameters; $K$-specific parameters follow the same initialization rule. The frozen K-aware diagnostic frame has teacher-residual effective rank $3.76$.

\begin{table}[ht]
\centering
\caption{GEO-vector count ablation under matched full-curriculum runs. Stage columns report the three-benchmark mean; recovery columns give individual benchmark scores. Throughput is measured during recovery.}
\label{tab:geo-count-full}
\scriptsize
\setlength{\tabcolsep}{3.1pt}
\begin{tabular*}{\linewidth}{@{\extracolsep{\fill}}rcccccccc@{}}
\toprule
$K$ & Joint & Bneck & SPAR & SPB & VS & Rec. mean & $r_{\mathrm{eff}}$ & Samples/s \\
\midrule
2 & 61.2 & 56.5 & 72.8 & 71.9 & 43.9 & 62.9 & 1.70 & 3.37 \\
\textbf{4} & \textbf{61.3} & 53.7 & \textbf{74.1} & 71.9 & \textbf{45.0} & \textbf{63.6} & 3.23 & 2.86 \\
8 & 60.4 & 55.9 & 72.8 & 72.0 & 42.5 & 62.4 & 3.81 & 2.46 \\
16 & 60.7 & 52.9 & 73.2 & \textbf{72.2} & 42.1 & 62.5 & \textbf{4.05} & 1.77 \\
\bottomrule
\end{tabular*}
\end{table}

All four settings remain non-collapsed. $K{=}2$ is task-effective but dimensionally constrained ($r_{\mathrm{eff}}{=}1.70$); $K{=}4$ approaches the teacher's residual capacity and gives the best recovery mean. Larger spans saturate the available structure without improving task performance, while recovery throughput falls from $2.86$ samples/s at $K{=}4$ to $1.77$ at $K{=}16$. This sweep suggests $K{=}4$ as a favorable compact operating point rather than a monotonic benefit from adding vectors. The main $K{=}8$ run was fixed for compatibility with the inherited GeoAnchor interface and belongs to a separate controlled run set, so the two $K{=}8$ endpoints are not directly comparable.

\subsection{Cross-backbone reproduction}
\label{app:cross-backbone}

We reproduce the joint--bottleneck--recovery pipeline with Qwen2.5-VL-3B. The two arms start from the same initialization, use the same 105{,}928 examples in the same batch order, and share all optimization and routing settings; only the GEO alignment loss differs. Both recovered endpoints use a learning rate of $10^{-5}$. Table~\ref{tab:qwen25-cross-backbone} shows that the structural distinction is reproduced on a second backbone: the model trained with GeoAnchor's GEO loss has a rank-one GEO span, whereas CR-GEO retains nearly four effective dimensions and nonzero teacher--GEO-vector assignment information. CR-GEO also improves both in-domain benchmarks and the three-benchmark mean, while ViewSpatial decreases by $1.70$ points.

\begin{table}[ht]
\centering
\caption{Matched cross-backbone reproduction with Qwen2.5-VL-3B. Benchmark entries are accuracy/score (\%); structure is measured on the same frozen 768-frame protocol used for the main backbone. Mean is the unweighted average of the three benchmark scores.}
\label{tab:qwen25-cross-backbone}
\small
\setlength{\tabcolsep}{4.2pt}
\begin{tabular}{lrrrrrr}
\toprule
GEO objective & SPAR & SPBench & ViewSpatial & Mean & $r_{\mathrm{eff}}$ & $I_{\mathrm{assign}}$ \\
\midrule
original $\mathcal{L}_{\mathrm{cov}}$ & 67.90 & 66.14 & \textbf{40.53} & 58.19 & 1.00 & $\approx0$ \\
\textbf{CR-GEO $\mathcal{L}_{\mathrm{CR\text{-}GEO}}$} & \textbf{68.63} & \textbf{67.83} & 38.83 & \textbf{58.43} & \textbf{3.84} & \textbf{0.178} \\
\bottomrule
\end{tabular}
\end{table}

Local POS/DIR decodability also transfers: CR-GEO obtains $0.396$\,m POS error and $0.888$ DIR cosine on the frozen held-out set. Replacing RGB with a different-scene image raises POS error from $0.444$ to $1.477$\,m and lowers DIR cosine from $0.922$ to $0.219$, confirming that the decoded local tokens depend on the corresponding input image.

\subsection{Additional diagnostics and qualitative visualizations}
\label{app:diagnostics}
\begin{figure}[!ht]
\centering
\begin{subfigure}{0.48\linewidth}
\centering
\includegraphics[width=\linewidth,trim=0 2pt 0 5pt,clip]{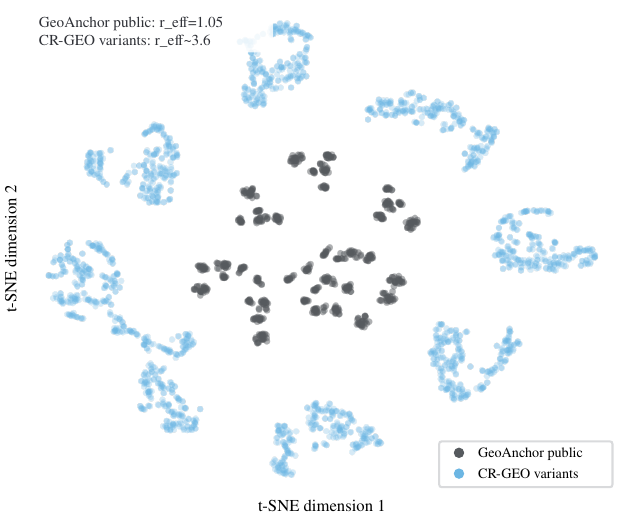}
\caption{GEO-latent t-SNE.}
\end{subfigure}\hfill
\begin{subfigure}{0.48\linewidth}
\centering
\includegraphics[width=\linewidth]{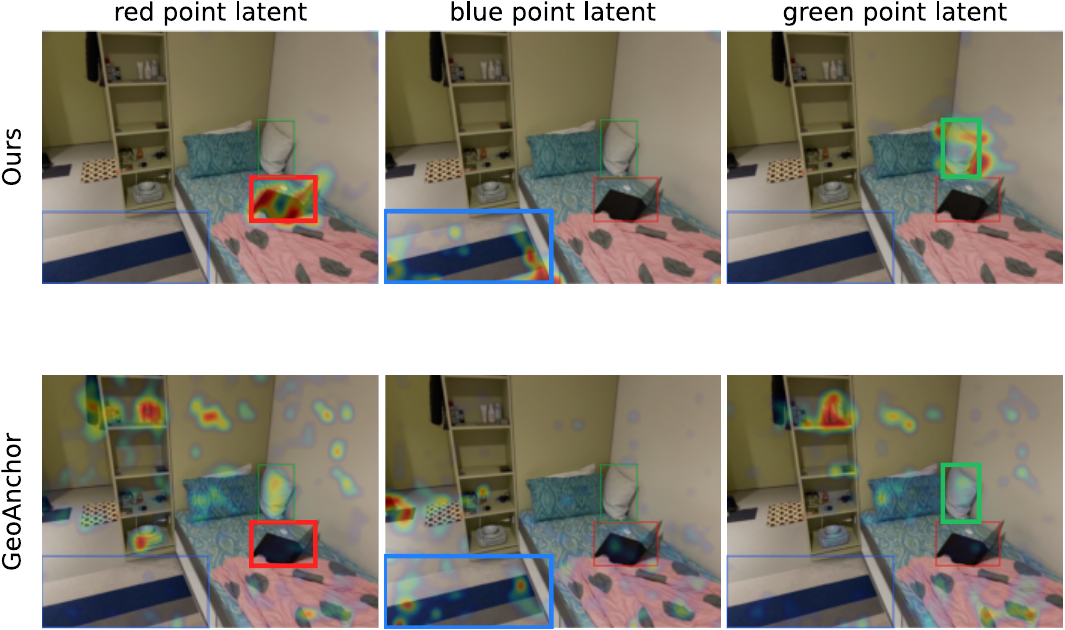}
\caption{Local-token attention.}
\end{subfigure}
\caption{Qualitative structure of the learned spatial latents. The t-SNE projection illustrates reduced concentration of GEO vectors under CR-GEO, and the local-token maps show object-specific visual attention during latent generation. Quantitative structure and use are reported in Table~\ref{tab:rep-appendix} and Figure~\ref{fig:kv}.}
\label{fig:appendix-qual}
\end{figure}

Figure~\ref{fig:appendix-qual} provides complementary qualitative views of latent spread and local-token attention.

\subsubsection{Local-token decoding across the curriculum}
\label{app:posdir}

We evaluate the six matched objective--stage checkpoints on 504 records from 248 scenes with zero scene, image, or record-ID overlap with the 105{,}928 training examples. The frame contains 1{,}134 POS and 79 DIR targets in the camera coordinate system. Table~\ref{tab:posdir-decoding} reports teacher-forced and live-generation measurements under each checkpoint's native routing mode.

\begin{table}[ht]
\centering
\caption{POS/DIR decoding across the matched $2{\times}3$ objective--stage grid. TF denotes teacher-forced evaluation. POS is mean per-point L2 error in meters; DIR and composed DIR are cosine similarities to ground truth. Live/replay DIR agreement is the cosine agreement between the DIR representation captured during live generation and the DIR representation obtained by teacher-forced replay of the same generated sequence. Live evaluation starts from the fixed 192-record subset; reported live metrics use the successfully aligned cases.}
\label{tab:posdir-decoding}
\scriptsize
\setlength{\tabcolsep}{3.4pt}
\resizebox{\linewidth}{!}{%
\begin{tabular}{llcccccc}
\toprule
Objective & Stage & TF POS $\downarrow$ & TF DIR $\uparrow$ & Composed DIR $\uparrow$ & Live POS $\downarrow$ & Live DIR $\uparrow$ & Live/replay DIR agreement $\uparrow$ \\
\midrule
CR-GEO & joint & 0.347 & 0.923 & 0.871 & 0.307 & 0.942 & 1.000 \\
CR-GEO & bottleneck & 0.365 & 0.911 & 0.864 & 0.345 & 0.919 & 0.971 \\
\textbf{CR-GEO} & \textbf{recovery} & \textbf{0.338} & 0.899 & \textbf{0.874} & \textbf{0.300} & 0.920 & \textbf{1.000} \\
original $\mathcal{L}_{\mathrm{cov}}$ & joint & 0.354 & 0.890 & 0.866 & 0.310 & 0.909 & 1.000 \\
original $\mathcal{L}_{\mathrm{cov}}$ & bottleneck & 0.376 & 0.876 & 0.862 & 0.345 & 0.884 & 0.957 \\
original $\mathcal{L}_{\mathrm{cov}}$ & recovery & 0.335 & 0.907 & 0.868 & 0.305 & 0.916 & 1.000 \\
\bottomrule
\end{tabular}%
}
\end{table}

Both objectives exhibit a temporary bottleneck-stage increase in POS error followed by full recovery. For CR-GEO, recovery changes POS by $-0.011$\,m relative to joint training (scene-bootstrap 95\% CI $[-0.020,-0.0001]$) and changes DIR cosine by $-0.024$ (CI $[-0.068,0.001]$). Across the six checkpoints, 176--185 of the 192 records yield aligned live measurements; among these successfully aligned cases, latent-span closure is 100\%. The final CR-GEO model also substantially exceeds train-derived constant priors (POS L2 $1.066$\,m; DIR cosine $0.113$). The POS prior is the mean of all training POS targets, and the DIR prior is the mean of all training DIR targets normalized for cosine evaluation; held-out evaluation data are not used to construct either prior. These comparisons establish POS/DIR decodability beyond a category or dataset mean.

For normalized teacher features $\hat{\mathbf v}_i$ and GEO vectors $\hat{\mathbf g}_j$, define the soft assignment
$p_{ij}=\mathrm{softmax}_{j}(\hat{\mathbf v}_i^\top\hat{\mathbf g}_j/\tau)$ with $\tau=0.07$ and its marginal $u_j=M^{-1}\sum_i p_{ij}$. We report
\begin{equation}
I_{\mathrm{assign}}=\frac{H(\mathbf u)-M^{-1}\sum_i H(\mathbf p_{i\cdot})}{\log K},
\end{equation}
which is zero when every teacher feature induces the same assignment distribution and increases when assignments are vector-specific and collectively distributed across the span. It is an assignment diagnostic, not a general neural mutual-information estimator.

We additionally report $H_{\mathrm{cond}}/\log K$, the normalized teacher-conditioned assignment entropy, and
\begin{equation}
N_{\mathrm{LSE}}
=
\operatorname{mean}_i
\exp\left(
\log\sum_j e^{A_{ij}}-\max_j A_{ij}
\right),
\end{equation}
which ranges approximately from $1$ to $K$ and measures the effective number of covering vectors.

\begin{table}[ht]
\centering
\caption{GEO-token representation diagnostics on frozen samples ($n{=}768$). $r_{\mathrm{eff}}$ is the effective rank of the $8{\times}8$ cosine Gram matrix. $I_{\mathrm{assign}}$ is normalized information induced by teacher-to-vector assignments, not a general neural mutual-information estimator. $N_{\mathrm{LSE}}$ is the effective number of covering vectors. Dashes denote diagnostics unavailable from historical captures, not failed measurements.}
\label{tab:rep-appendix}
\small
\begin{tabular}{lccccc}
\toprule
Model & off-diag cos & $r_{\mathrm{eff}}$ & $H_{\mathrm{cond}}/\log K$ & $I_{\mathrm{assign}}$ & $N_{\mathrm{LSE}}$ \\
\midrule
original $\mathcal{L}_{\mathrm{cov}}$, matched-init joint & 1.000 & 1.00 & -- & $\approx0$ & -- \\
CR-GEO, matched-init joint & 0.167 & 3.87 & -- & 0.228 & -- \\
GeoAnchor public & 0.991 & 1.05 & 0.999 & $\approx0$ & 7.54 \\
original $\mathcal{L}_{\mathrm{cov}}$, joint (reference weights) & 0.460 & 2.73 & 0.667 & $\approx0$ & 4.00 \\
original $\mathcal{L}_{\mathrm{cov}}$, recovery (reference weights) & 1.000 & 1.00 & 1.000 & $\approx0$ & 7.99 \\
original $\mathcal{L}_{\mathrm{cov}}$, joint (GeoLatent weights) & 1.000 & 1.00 & -- & $\approx0$ & -- \\
original $\mathcal{L}_{\mathrm{cov}}$, recovery (GeoLatent weights) & 1.000 & 1.00 & -- & $\approx0$ & -- \\
CR-GEO, main-curriculum joint & -- & 2.72 & -- & 0.048 & -- \\
CR-GEO, direct NTP continuation & 0.425 & 2.85 & 0.803 & 0.043 & 3.69 \\
CR-GEO, bottleneck $\rightarrow$ NTP-only & 0.216 & 3.55 & 0.649 & 0.055 & 2.43 \\
GeoLatent, recovery & \textbf{0.224} & \textbf{3.62} & \textbf{0.579} & \textbf{0.071} & \textbf{2.00} \\
\bottomrule
\end{tabular}
\end{table}

\subsubsection{Decomposed-latent route interventions}
\label{app:carrier-cuts}
We apply attention-edge interventions to independently trained bottleneck models with four or eight GEO vectors on the same frozen 64 scenes (128 directions). ``Block P/D'' prevents subsequent non-latent queries from reading POS and DIR key/value slots; ``Block latent readout'' additionally blocks downstream non-latent readout of GEO keys while preserving latent generation and latent-to-latent relay. Token identities, projector outputs, images, and the prefix preceding each intervention are unchanged. All conditions use a fixed denominator, and mask-difference checks verify that only the intended edges change.

\begin{table}[ht]
\centering
\caption{Decomposed-latent localization under bottleneck inference.
Entries are accuracy (drop from the no-cut condition), in percent.
``Block prior-latent access'' prevents all subsequent typed and non-typed
queries from attending to previously generated typed-latent keys.
For both $K{=}4$ and $K{=}8$, neither the POS+DIR drop nor the
latent-readout drop was matched or exceeded by any of 20{,}000
two-sided scene-clustered permutations (plus-one $p{=}1/20001$ for each).}
\label{tab:carrier-cuts}
\scriptsize
\setlength{\tabcolsep}{4pt}
\begin{tabular*}{\linewidth}{@{\extracolsep{\fill}}lrrrr@{}}
\toprule
& \multicolumn{2}{c}{$K{=}4$} & \multicolumn{2}{c}{$K{=}8$} \\
\cmidrule(lr){2-3}\cmidrule(lr){4-5}
Condition & Acc. & Drop & Acc. & Drop \\
\midrule

No latent-readout cut & 85.16 & -- & 89.06 & -- \\
Block POS & 61.72 & 23.44 & 53.12 & 35.94 \\
Block DIR & 81.25 & 3.91 & 85.94 & 3.13 \\
Block POS+DIR & 26.56 & \textbf{58.59} & 26.56 & \textbf{62.50} \\
Block GEO & 85.16 & 0.00 & 89.06 & 0.00 \\
Block latent readout & 27.34 & \textbf{57.81} & 25.78 & \textbf{63.28} \\
Block prior-latent access & 22.66 & \textbf{62.50} & 27.34 & \textbf{61.72} \\
\bottomrule
\end{tabular*}
\end{table}

On the same frozen pairs and under the same four-dimensional additive attention-mask protocol (batch, head, query, key), the public GeoAnchor checkpoint remains at $82.03\%$ before and after the latent-readout cut. Replication across $K$ supports the stage-level mechanism on this frozen direction subset: once the direct visual path is closed, answers depend on the decomposed latent intermediates. On this direction subset, jointly cutting POS and DIR removes nearly all above-chance accuracy. This dependence on the latent intermediates is the targeted effect of bottleneck training.

\begin{table}[ht]
\centering
\caption{Stage-wise route intervention in the matched $K{=}8$ trajectory on the same 128 directions. ``Block latent readout'' hides all decomposed-latent keys from subsequent non-latent queries while preserving latent-to-latent relay.}
\label{tab:stage-route}
\scriptsize
\setlength{\tabcolsep}{8pt}
\begin{tabular}{lrrr}
\toprule
Checkpoint & No latent-readout cut & Block latent readout & Drop \\
\midrule
Joint (full attention) & 69.53 & 68.75 & 0.78 \\
Bottleneck (routed) & 89.06 & 25.78 & \textbf{63.28} \\
Recovery (full attention) & 82.03 & 81.25 & 0.78 \\
\bottomrule
\end{tabular}
\end{table}

The intervention changes accuracy by only $0.78$ points at joint, by $63.28$ points at bottleneck, and again by $0.78$ points after recovery. This profile follows the joint $\rightarrow$ bottleneck $\rightarrow$ recovery sequence: bottleneck training routes image evidence through the decomposed latents during answer generation, whereas recovery restores direct image access. The intervention therefore captures stage-specific dependence on the latent intermediates induced by bottleneck training, rather than a permanent requirement after recovery.

We next test whether recovery preserves a usable latent-mediated visual route over each complete benchmark. All conditions use the same intervention-compatible attention path. In \emph{latent-only}, non-latent queries cannot read image keys, while latent-token queries retain image access and downstream queries retain latent-token access. In \emph{neither}, downstream queries additionally cannot read latent-token keys. Their paired difference therefore isolates the image$\rightarrow$latent$\rightarrow$downstream route without requiring it to be the model's only route under ordinary full attention. Table~\ref{tab:route-evidence}(b) reports the corresponding full-benchmark accuracies in the main text.

GeoLatent's latent-only route retains $71.3\%$, $71.1\%$, and $77.4\%$ of its full-attention score and outperforms the corresponding neither condition by $5.31$, $3.72$, and $1.76$ points on SPAR-Bench, SPBench, and ViewSpatial, with clustered 95\% CIs $[1.81,9.03]$, $[2.16,5.31]$, and $[0.53,5.98]$. The direct-recovery control gains approximately zero. The curriculum difference-in-differences estimates are $5.37$ points (95\% CI $[1.78,9.10]$, $p{=}0.0313$), $4.06$ points (95\% CI $[2.09,5.96]$, $p{=}0.00030$), and $1.78$ points (95\% CI $[0.47,6.33]$, plus-one $p{=}1/20001$), respectively. Resampling and sign flips use the predefined task clusters for SPAR-Bench and scene clusters for SPBench/ViewSpatial. Thus bottleneck training establishes a latent-mediated capability that remains available after recovery, even though ordinary full-attention inference also has a restored direct visual route.

Together, the diagnostics separate stage-specific evidence. Rank and assignment statistics measure global GEO organization, while POS/DIR decoding measures local geometric decodability. At the bottleneck checkpoint, edge interventions show use of the latent intermediates; after recovery, the decomposed representation remains structured while direct image access is restored.

\FloatBarrier
\subsection{Controlled no-GEO capability profile}
\label{app:nogeo}

The same-data no-GEO control follows the full curriculum on all 105{,}928 samples while removing the GEO span and its VGGT supervision. Its $62.4\%$ three-benchmark mean falls to $40.3/38.1\%$ under blank/mismatched images, confirming substantial dependence on visual input. GeoLatent reaches $62.8\%$ while additionally providing a differentiated GEO span ($r_{\mathrm{eff}}{=}3.62$, $I_{\mathrm{assign}}{=}0.071$), stronger ViewSpatial visual reliance (VEF $88.3\%$ versus $60.1\%$), and a $5.9$-point camera-relative-direction gain. Person-centered direction additionally requires observer orientation $R_{\mathrm{person}}$, which the current camera-frame vocabulary does not encode and which motivates future extensions.

\end{document}